\documentclass[conference]{IEEEtran}
\usepackage{cite}

\ifCLASSINFOpdf
\else
\fi
\usepackage{url}
\PassOptionsToPackage{hyphens}{url}

\usepackage{makecell}
\usepackage{multirow}
\usepackage{amsmath,amssymb,amsfonts}
\usepackage{graphicx} 
\graphicspath{ {images/} }
\usepackage[font=normalsize]{subfig}
\usepackage{algorithm}
\usepackage{algpseudocode}
\usepackage{xcolor} 

\algnewcommand{\LineComment}[1]{\State \(\triangleright\) #1}

\begin{document}
%
\title{Promoting High Diversity Ensemble Learning with EnsembleBench}


\author{\IEEEauthorblockN{Yanzhao Wu, Ling Liu, Zhongwei Xie, Juhyun Bae, Ka-Ho Chow, Wenqi Wei}
\IEEEauthorblockA{School of Computer Science, 
Georgia Institute of Technology\\
Email: yanzhaowu@gatech.edu, lingliu@cc.gatech.edu, \{zhongweixie,juhyun.bae,khchow,wenqiwei\}@gatech.edu}
}


%


\maketitle

\begin{abstract}
Ensemble learning is gaining renewed interests in recent years. This paper presents EnsembleBench, a holistic framework for evaluating and recommending high diversity and high accuracy ensembles. The design of EnsembleBench offers three novel features: (1) EnsembleBench introduces a set of quantitative metrics for assessing the quality of ensembles and for comparing alternative ensembles constructed for the same learning tasks. (2) EnsembleBench implements a suite of baseline diversity metrics and optimized diversity metrics for identifying and selecting ensembles with high diversity and high quality, making it an effective framework for benchmarking, evaluating and recommending high diversity model ensembles. (3) Four representative ensemble consensus methods are provided in the first release of EnsembleBench, enabling empirical study on the impact of consensus methods on ensemble accuracy. A comprehensive experimental evaluation on popular benchmark datasets demonstrates the utility and effectiveness of EnsembleBench for promoting high diversity ensembles and boosting the overall performance of selected ensembles.
\end{abstract}


%
\IEEEpeerreviewmaketitle

\section{Introduction}
\noindent 
Ensemble learning harnesses the combined and complementary wisdom of multiple base models (learners) to improve the prediction performance over individual models. High diversity ensembles have high failure independence and low negative correlation, which lead to high ensemble accuracy. Although ensemble learning research has been around for three decades, many questions remain unanswered regarding (1) the quantitative properties of a good ensemble and (2) criteria for choosing a good ensemble from a pool of $M$ individual models.

\subsection{Related Work and Problem Statement}
\subsubsection{Homogeneous Ensembles} 
Early efforts on ensemble learning are represented by bagging~\cite{bagging}, boosting~\cite{boosting} and random forests~\cite{randomforests}. Given a user-defined parameter $M$, these approaches all use iterative training with the same training algorithm and the same loss optimization to produce the $M$ member models of the ensemble. Bagging-based ensemble uses random sampling with error-margin controlled replacement to build the training set for each round in the iterative training process, such that training samples with large error margin will be sampled more times in producing the next version of the training set. Boosting-based ensemble uses the error-margin to derive the model weight update such that training samples with larger error margin in the current iteration will receive larger weight in the next iteration through the boosting algorithm. For random forest, it builds $M$ decision trees iteratively by randomly selecting attributes to serve as the nodes of the $M$ trees and use information gain to revise the node split decision iteratively by minimizing error margin. We call the ensembles generated by iterative training algorithms the homogeneous ensembles, because they share the same training algorithm with the same loss function and loss optimization. Generally speaking, given the parameter $M$, a homogeneous ensemble of $M$ member models will be produced by an iterative training algorithm, such as Boosting, Bagging or Random Forest. Random Forest is the most representative and has been successfully deployed in many classification learning applications~\cite{randomforests, randomdecisionforests}. 

\subsubsection{Q-Diversity Metrics}
The popularity of random forest has triggered research studies on diversity metrics, which attempt to provide diversity as a quantitative measure for explaining and understanding of ensemble learning. The representative \textbf{pairwise} diversity metrics include Cohen's Kappa (CK)~\cite{cohenskappa}, Q Statistics (QS)~\cite{qstatistics}, Binary Disagreement (BD)~\cite{binarydisagreement}. The representative \textbf{non-pairwise} diversity metrics include Fleiss' Kappa (FK)~\cite{fleisskappa}, Kohavi-Wolpert Variance (KW)~\cite{kwvariance,diversityaccuracy} and Generalized Diversity (GD)~\cite{generalizeddiversity}. We coin these baseline diversity metrics as Q-diversity metrics in our EnsembleBench. Several studies indicate the correlation between diversity metrics and ensemble accuracy~\cite{diversityaccuracy, generalizeddiversity}. However, given a pool of $M$ base models, these Q-diversity metrics are primarily used to compare and evaluate ensemble diversity and ensemble accuracy for only ensembles that are composed of all $M$ base models. Few addresses the question of whether a subset of the $M$ base models as an ensemble could outperform the ensemble of all $M$ models.  

\subsubsection{Heterogeneous ensembles} 
Heterogeneous ensembles typically refer to those ensembles that are constructed by random teaming from the pool of $M$ pre-trained models. For example, a pool of $M$ pre-trained models may include linear SVM, RBF SVM, Decision Tree, na\"ive bayes, Gaussian Process, Neural Network models trained using different backbone algorithms, such as ResNet, AlexNet, VGG, DenseNet. For benchmark datasets, such as CIFAR-10, ImageNet, one can obtain such pre-trained models from well-known public model zoos~\cite{caffe, onnxmodelzoo, gtmodelzoo}. There are several open questions for heterogeneous ensembles: (1) Does ensemble with more member models outperform ensemble with fewer member models? (2) Given a pool of $M$ base models, are there more than one good quality ensembles and how to find them? (3) Can the Q-diversity metrics be useful for identifying and selecting high performance ensembles? And are they effective? (4) What quantitative metrics are informative measures for evaluating and comparing the quality of alternative ensembles, given a learning task? 
Regarding the question (1), our existing research on mitigating adversarial attacks has exploited model ensemble learning to improve adversarial robustness of the victim model~\cite{ensemble-mass,ensemble-bigdata,robustdlensemble}. Through this ensemble mitigation research, we learned two interesting facts: First, ensembles of high Kappa~\cite{cohenskappa} diversity can provide improved robustness against adversarial perturbed examples and out-of-distribution samples~\cite{robustdlensemble}. Second, ensembles of smaller size (fewer member models) can outperform the ensembles of larger size (with more member models) in adversarial learning context. However, we also experienced several frustrations. Among the set of candidate ensembles, even for $M=5$, it is non-trivial to pick out the best performing ensembles. The ensemble with best Kappa diversity sometimes may not outperform a randomly picked ensemble team. This prior experience and our efforts to address these questions have led us to develop EnsembleBench.

\subsection{Scope and Contribution of this paper}
\noindent In this paper, we present EnsembleBench, a holistic framework for evaluating and recommending high accuracy ensembles by promoting high diversity ensemble learning. EnsembleBench is original in three aspects: (1) EnsembleBench introduces a set of quantitative metrics for assessing the quality of ensembles and for comparing alternative ensembles constructed for the same learning tasks. (2) EnsembleBench implements a suite of baseline diversity metrics and optimized diversity metrics for identifying and selecting ensembles with high diversity and high quality, making it an effective framework for benchmarking, evaluating and recommending high diversity model ensembles. (2) EnsembleBench provides a suite of representative ensemble consensus methods, such as model averaging, majority voting, plurality voting, and learn to combine. Our empirical study shows that these consensus methods have similar performance impact on ensemble accuracy of the selected ensemble teams for the benchmark datasets. We report our experimental evaluation on two popular benchmark datasets: CIFAR-10~\cite{cifar10-100} and ImageNet~\cite{ILSVRC}. We show the utility and effectiveness of EnsembleBench for promoting high diversity ensembles and boosting the overall performance of selected ensembles. We also show that the optimized ensemble diversity metrics and algorithms for ensemble selection can significantly improve the prediction performance of selected ensembles. We would like to note that although this paper focuses on ensembles of pre-trained models (heterogeneous ensembles), the evaluation framework and metrics are generally applicable to evaluation and comparison of homogeneous ensembles generated by iterative training algorithms~\cite{bagging, boosting, randomforests}.

\begin{figure*}[h!]
\centering
    \includegraphics[width=1.0\textwidth]{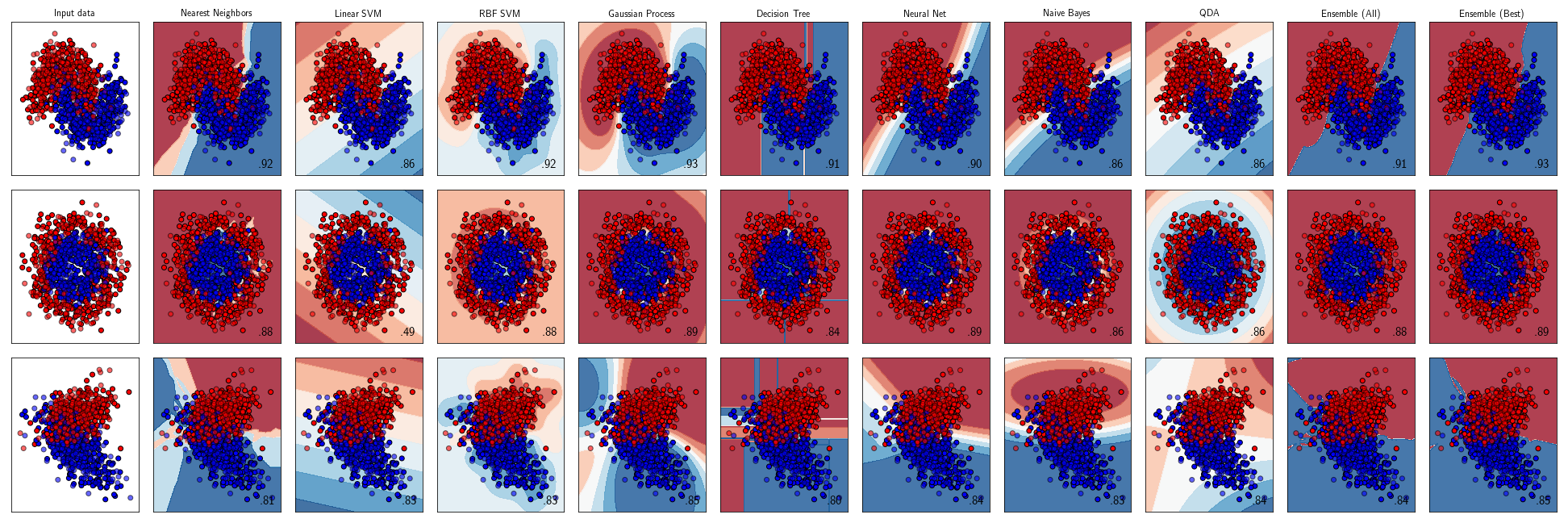}
    \caption{Prediction Performance Visualization of Individual Base Models and Ensembles}
    \label{fig:visualization-ensemblebench}
\end{figure*}

\begin{table*}[h!]
\centering
\caption{Prediction Accuracy (\%) Comparison}
\label{table:prediction-acc-comparison}
\scalebox{0.9}{
\small
\begin{tabular}{|c|c|c|c|c|c|c|c|c|c|c|c|c|}
\hline
Model & \multirow{2}{*}{0} & \multirow{2}{*}{1} & \multirow{2}{*}{2} & \multirow{2}{*}{3} & \multirow{2}{*}{4} & \multirow{2}{*}{5} & \multirow{2}{*}{6} & \multirow{2}{*}{7} & \multicolumn{4}{c|}{Ensemble} \\ \cline{1-1} \cline{10-13} 
Dataset &  &  &  &  &  &  &  &  & All & Best & 026 & 045 \\ \hline
0 & 91.50 & 86.25 & 91.75 & \textbf{92.50} & 90.50 & 91.00 & 85.75 & 86.00 & 92.00 & \textbf{92.75} & 92.50 & 91.75 \\ \hline
1 & 87.50 & 48.75 & 88.00 & \textbf{88.50} & 84.00 & 88.25 & 86.50 & 86.50 & 87.25 & \textbf{89.25} & 88.75 & 89.00 \\ \hline
2 & 81.25 & 83.50 & 83.25 & \textbf{84.75} & 80.75 & 83.50 & 83.25 & 84.25 & 84.50 & \textbf{85.00} & 83.00 & 81.75 \\ \hline
\end{tabular}
} 
\end{table*}

\begin{table*}[h!]
\centering
\caption{Ensemble Evaluation}
\label{table:ensemble-evaluation}
\scalebox{0.8}{
\small
\begin{tabular}{|c|c|c|c|c|c|c|c|c|c|c|}
\hline
Dataset & \#EnsSet & \begin{tabular}[c]{@{}c@{}}Ensemble Acc\\ Range (\%)\end{tabular} & \begin{tabular}[c]{@{}c@{}}Ensemble\\ Acc Avg (\%)\end{tabular} & STD & \begin{tabular}[c]{@{}c@{}}\# (Acc \textgreater{}=\\ m\_max)\end{tabular} & \begin{tabular}[c]{@{}c@{}}\% (Acc \textgreater{}= \\ m\_max)\end{tabular} & \begin{tabular}[c]{@{}c@{}}Max Pool Model\\ Acc (\%, p\_max)\end{tabular} & \begin{tabular}[c]{@{}c@{}}\# (Acc \\ \textgreater{}= p\_max)\end{tabular} & \begin{tabular}[c]{@{}c@{}}\% (Acc \\ \textgreater{}= p\_max)\end{tabular} & Best Ensemble \\ \hline
0 & 247 & 85.75$\sim$92.75 & 90.35 & 2.23 & 62 & 25.10\% & 92.50 & 16 & 6.48\% & 92.75\% ({[}0, 1, 2, 3{]}) \\ \hline
1 & 247 & 48.75$\sim$89.25 & 86.02 & 6.45 & 36 & 14.57\% & 88.50 & 23 & 9.31\% & 89.25\% ({[}0, 3, 4{]}) \\ \hline
2 & 247 & 81.75$\sim$85.00 & 83.95 & 0.58 & 71 & 28.74\% & 84.75 & 22 & 8.91\% & 84.50\% ({[}3, 4, 6{]}) \\ \hline
\end{tabular}
} 
\end{table*}

\section{Ensemble Teaming and Evaluation} \label{section:ensemble-evaluation}
\subsection{Baseline Teaming for Candidate Ensembles}
\noindent Consider a pool of $M$ base models for a learning task and dataset $\mathcal{D}$, denoted by $BMSet(\mathcal{D})$= $\{F_0, ..., F_{M-1}\}$, let $EnsSet$ denote the set of candidate ensembles that are formed from the base model pool $BMSet(\mathcal{D})$, and $S$ denote the ensemble team size, ranging from 2 to $M$, then the number of candidate ensembles is $\binom{M}{S}$, and $|EnsSet|=\sum_{S=2}^{M}\binom{M}{S} = \binom{M}{2} + \binom{M}{3} + ... + \binom{M}{M} = 2^M - (1+M)$. The size of the candidate ensemble set $BMSet(\mathcal{D})$ grows exponentially with the number of base models ($M$). For example, with $M=8$, we have $|EnsSet| = 247$. For $M=10$, $|EnsSet|=1,013$ and for $M=15$,  $|EnsSet|=32,752$. Hence, as $M$ increases, it is a non-trivial problem to select high performance ensembles from the set of candidate ensembles in $EnsSet$.

Figure~\ref{fig:visualization-ensemblebench} provides a visualization generated from scikit-learn~\cite{sklearn-classifier-comparison} for $M=8$ individual base models on three datasets (three rows) for binary classification. The last two columns show the two ensemble teams, one with all 8 models and the other is the best performing ensemble for each of the three datasets. The plurality voting is used.
Table~\ref{table:prediction-acc-comparison} shows the test accuracy for all 8 models and the four example ensembles on the three datasets (0$\sim$2). The 8 base models are in the second to the ninth columns and they are KNN, Linear SVM, RBF SVM, Gaussian Process, Decision Tree, Neural Net, Na\"ive Bayes and QDA (Quadratic Discriminant Analysis), denoted by $F_i$ ($0\leq i\leq 7$) respectively. The last four columns are the ensemble with all 8 base models (All), the ensemble with the highest prediction accuracy (Best) identified by EnsembleBench and two example ensembles $F_0 F_2 F_6$ and $F_0 F_4 F_5$. The best ensembles for the three datasets are $F_0 F_1 F_2 F_3$, $F_0 F_3 F_4$ and $F_3 F_4 F_6$ as shown on the last column of Table~\ref{table:ensemble-evaluation} respectively.

We highlight three interesting observations.
{\it First}, the best ensemble outperforms any member models in the entire pool of $M$ models: 92.75\% \textgreater{} 92.50\% on Dataset 0, 89.25\% \textgreater{} 88.50\% on Dataset 1 and 85.00\% \textgreater{} 84.75\% on Dataset 2. Note that $F_3$ (Gaussian Process) has the highest accuracy among all $M$ base models for all three datasets but the best ensemble for all three datasets.
{\it Second}, the best ensembles for all three datasets outperform the ensemble with all $M=8$ models. Also some member models in the best ensemble are not among the top 5 accuracy ranked models. For instance, on Dataset 0, the base model $F_1$ (only 86.25\% accuracy) is included in the best ensemble $F_0 F_1 F_2 F_3$. 
{\it Third}, ensembles without the base model with the highest accuracy (max) in the base model pool may outperform the member model with the max accuracy in the pool. For example, the base model $F_3$ has the max accuracy 88.50\% on Dataset 1, and the two ensembles: $F_0 F_2 F_6$ and $F_0 F_4 F_5$, identified by EnsembleBench, achieved 88.75\% and 89.00\% accuracy respectively. 
These observations further confirm that (1) ensembles of base models hold the potential to improve the prediction performance of individual member models, (2) the ensemble of all $M$ base models may not be the top performing ensemble, and may not even outperform the member model with the max accuracy (recall Ensemble (All) and $F_3$ for the three datasets), (3) algorithms and metrics that can select the top performing ensembles from the large candidate ensemble set are critical, and (4) quantitative metrics to evaluate ensemble quality are equally important. 

\subsection{Evaluation Metrics for High Quality Ensembles}
\noindent Recall examples in Figure~\ref{fig:visualization-ensemblebench} and Table~\ref{table:prediction-acc-comparison}, we can see that ensemble learning has an absolute gain if the ensemble learning improves the prediction performance over every of its member models. We consider this criterion as the first ensemble quality measure. Given the context of the base model pool of $M$ models, one may also set the goal of searching for the high quality ensemble teams that not only outperform every of its member models but also improve over every individual model in the entire base model pool. Clearly, this second criterion is unnecessarily strong although such ensembles do exist. In EnsembleBench, an ensemble is absolutely high quality, if it satisfies the first criterion. We use the second criterion to compare and rank the set of superior ensembles. Furthermore, we also introduce the following quantitative measures to compare two different ensemble selection algorithms: \textit{ensemble accuracy range}, \textit{average ensemble accuracy}, and \textit{standard deviation (STD)} of ensemble accuracy, all measured over the set of ensembles selected by an ensemble selection/recommendation algorithm in EnsembleBench.

Consider the same example in Figure~\ref{fig:visualization-ensemblebench} and the baseline ensemble teaming method, for $M=8$, we have $|EnsSet|=2^8 - (8+1)=247$, i.e., a total of 247 candidate ensembles. Table~\ref{table:ensemble-evaluation} shows the evaluation of candidate ensembles on the three datasets. For the 247 candidate ensembles, we compute the ensemble accuracy range, the average ensemble accuracy and the standard derivation. Next we also compute the number of ensembles with ensemble accuracy higher than the max accuracy of its member models (Acc \textgreater{}= m\_max) and the percentage of such ensembles. In addition, we use the max accuracy of the individual model in the base model pool as the reference point, and compute the number and the percentage of ensembles that outperform the base model with the max accuracy over the pool of $M$ models (Acc \textgreater{}= p\_max) on the 6th$\sim$10th columns in Table~\ref{table:ensemble-evaluation}.
The final column lists the best ensemble and its ensemble accuracy. 
We made two observations.
{\it First}, the average accuracy of all candidate ensembles is high, with 90.35\%, 86.02\% and 84.75\% for three datasets respectively. Meanwhile, their STD is relatively low, ranging from 0.58 to 6.45. This indicates that a large number of ensemble teams can provide high ensemble accuracy. A randomly selected ensemble could offer the performance on par to the average ensemble accuracy.
{\it Second}, the probability of selecting an ensemble with absolutely high quality from the set of 247 candidate ensembles is relatively low. For all three datasets, only 14.57\% $\sim$ 28.74\% of candidate ensembles can outperform their respective member models. Moreover, there are a small percentage of ensembles with superior quality, indicating that they outperform the individual model with the max accuracy in the base model pool, i.e., 6.48\% $\sim$ 9.31\% for the three datasets. These results motivate us to further explore algorithms that can identify 
and recommend high quality ensembles and prune out low quality ensembles.
In the subsequent sections, we first give an overview of EnsembleBench design and then focus on Q-diversity-based ensemble selection algorithms, their inherent problems, optimizations and new diversity metrics introduced to improve the performance of ensemble selection algorithms.

\section{EnsembleBench}
\begin{figure}[h!]
\centering
    \includegraphics[width=0.5\textwidth]{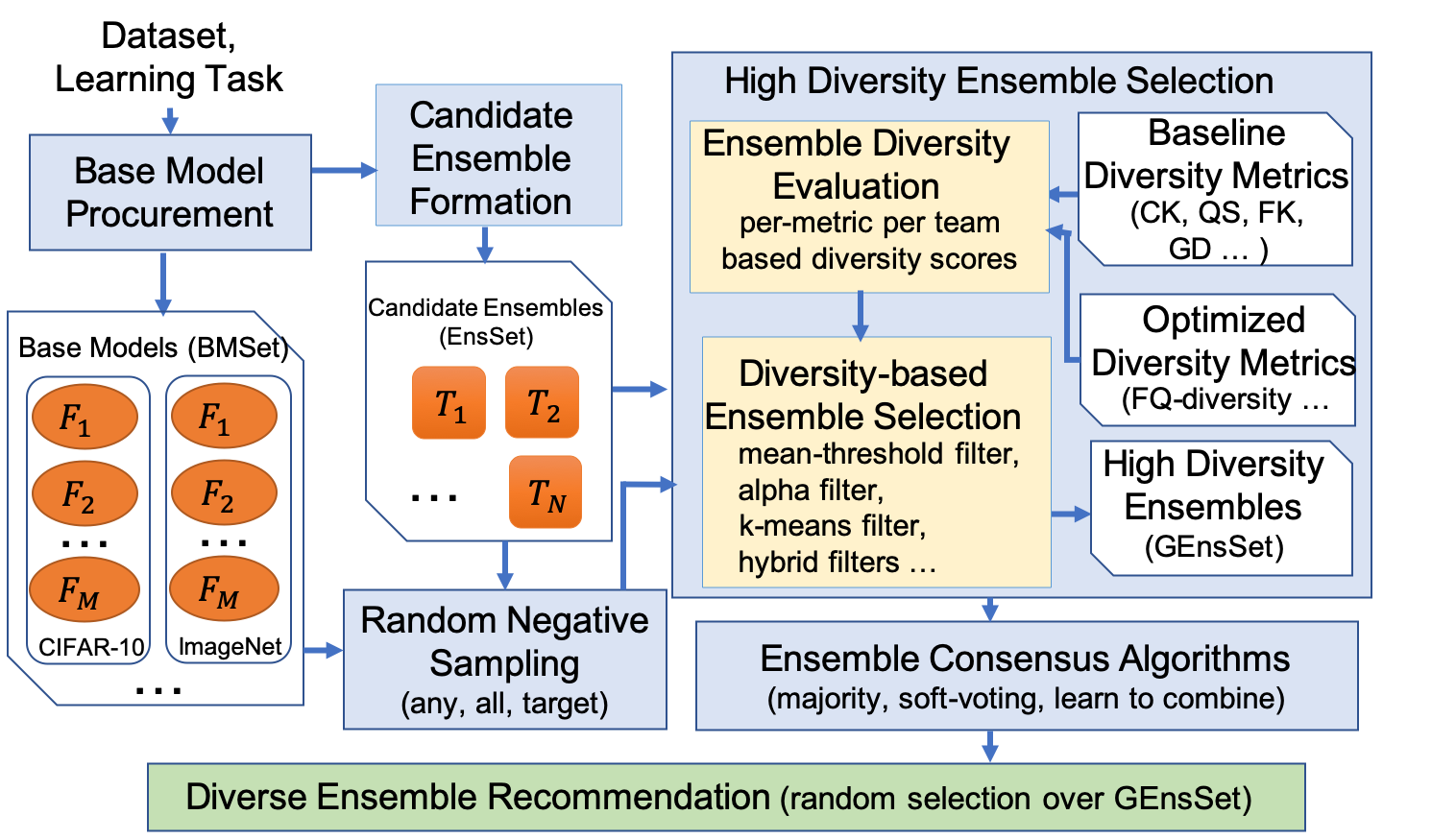}
    \caption{EnsembleBench Architecture}
    \label{fig:ensemblebench-overview}
\end{figure}

\noindent
EnsembleBench is a holistic framework for evaluating and recommending high performance ensembles by promoting ensemble diversity and boosting ensemble accuracy. It consists of seven functional components: (1) base model procurement, (2) candidate ensemble formation, (3) random negative sampling, (4) diversity-based ensemble evaluation, (5) diversity-based ensemble selection, (6) ensemble consensus algorithms, and (7) diverse ensemble recommendation. Figure~\ref{fig:ensemblebench-overview} provides an architectural overview of our EnsembleBench system. Although the public model zoos~\cite{caffe,onnxmodelzoo,GTDLBenchTSC,GTDLBenchBigData,GTDLBenchICDCS} are a primary source for the base model procurement, contributions from privately trained models~\cite{ensemble-bigdata, ensemble-icnc, LRBenchBigData} are also collected. The first open-source release of EnsembleBench is available on GitHub (\url{https://github.com/git-disl/EnsembleBench}).

\subsection{Workflow of EnsembleBench}
\noindent Given a learning task and training dataset, a user of EnsembleBench can choose a pool of $M$ base models from the database of base models, which is managed by the base model procurement module and organized by the learning task and training dataset. Then, the candidate ensemble formation module will be invoked to produce the set of candidate ensembles ($EnsSet$). The next step is to invoke diversity selection algorithms to obtain the set of ensembles recommended by the diversity based ensemble evaluation module. For each diversity metric, the diversity score will be calculated based on negative samples generated from the random negative sampling module. The diversity based ensemble selection module will evaluate every candidate ensembles in $EnsSet$ and output the set of selected ensembles ($GEnsSet$), which are high quality ensembles according to each diversity metric. Different diversity metrics result in different ensemble selection algorithms, which in turn produce different sets of ensemble recommendations due to the variations in defining ensemble diversity. Such diversity score is intended to reflect the degree and complexity of the abstract correlation between ensemble diversity and ensemble accuracy in the hope that there is high probability for ensembles with high diversity to result in high ensemble accuracy. The ensemble consensus module will compute the ensemble accuracy for each selected ensemble team using one of the consensus algorithms, such as majority voting and soft voting (model averaging), to obtain the ensemble prediction results. Hence, in addition to compute Q-diversity scores for each ensemble in the candidate set ($EnsSet$), EnsembleBench can be used as a benchmarking tool to evaluate and compare the quality of the selected ensembles produced by using different ensemble diversity metrics and their corresponding ensemble selection algorithms. Furthermore, for a set of recommendations, multiple statistical measurements are provided with illustrative visualization of the criteria used by the recommended selection algorithm. The users may choose their most preferred recommendation based on multiple quantitative measures of the selected ensembles and the corresponding diversity scores, of which the ensemble selection decision is made.
There are several ways to utilize the set of recommended ensembles in $GEnsSet$. For example, both a random ensemble selection over $GEnsSet$ and a random selection over the subset of $GEnsSet$ filtered by a diversity threshold, have been used in model ensemble approaches for mitigating adversarial inputs, such as MODEF~\cite{ensemble-bigdata} and XEnsemble~\cite{ensemble-icnc, robustdlensemble}.

\subsection{Use-cases of EnsembleBench}
\noindent In addition to the above first use case, the second use case could be to ask EnsembleBench whether a specific ensemble of $K$ member models is a good quality ensemble in terms of ensemble diversity. We consider three cases: (1) If EnsembleBench has an existing base model pool containing all $K$ models in the query, then a search on the querying ensemble team of $K$ member models will return the recommendation with a set of explainable and quantitative measures to illustrate the logic used for producing the recommendation. 
Such recommendation should include the diversity score and negative sample based ensemble accuracy for the ensemble of all $K$ models as well as those ensembles which consist of only some proper subset of $K$ models and yet offer high ensemble diversity and high ensemble accuracy measured using negative samples generated from the random negative sampling module. The quantitative measures also include average accuracy and standard derivation and the accuracy lower and upper bounds for the selected set of ensembles using the $K$ querying models as the base model pool. Recall examples in Figure~\ref{fig:visualization-ensemblebench}, Table~\ref{table:prediction-acc-comparison} and~\ref{table:ensemble-evaluation}, one may ask if the ensemble of three member models (045) is high quality ensemble for the Dataset 1, the result in the row with Dataset 1 will provide a clear explanation. 
(2) If EnsembleBench has an existing base model pool containing a majority of the $K$ models, e.g., only one model in the ensemble team of the $K$ member models is new to EnsembleBench, then EnsembleBench will create a base model pool with this $K$ models. Then perform incremental evaluation by computing the diversity scores for candidate ensembles that contain this new base model. Then the same workflow as that in case (1) will be executed to produce the answer.
(3) If all $K$ models are new, then EnsembleBench will perform the entire workflow of evaluation on this new pool of $K$ models and produce the recommendation with explainable and quantitative measures as outlined in the previous section. 

\section{Q-Diversity based Ensemble Selection}
\noindent 
Diversity metrics are classified into two broad categories based on whether the diversity of an ensemble is computed by pair-wise diversity and then taking the average or by using non-pairwise diversity formula. In the first release of EnsembleBench, we implement three representative pairwise Q-diversity metrics: Cohen's Kappa (CK), Q Statistics (QS) and Binary Disagreement (BD), and three non-pairwise metrics: Fleiss' Kappa (FK), Kohavi-Wolpert Variance (KW) and Generalized Diversity (GD). An ensemble selection algorithm has two functionalities: providing guidelines that can be used to choose high performance ensembles and pruning out low quality ensembles based on a given diversity metric. All six Q-diversity metrics were originally introduced to analyze the correlation between ensemble diversity and ensemble accuracy. In EnsembleBench, we implement Q-diversity based ensemble selection algorithms with dual goals. First, we want to analyze the effectiveness of using these Q-diversity metrics for ensemble selection. Second, these Q-diversity based ensemble selection algorithms can be used as diversity ensemble selection baselines. 

Given an ensemble of $S$ member models on a training set $\mathcal{D}$, all six Q-diversity metrics are computed by using a set of negative samples randomly selected from the collection of negative samples based on the $S$ member models and the training set $\mathcal{D}$. Let $\mathcal{D} = \{\textbf{x}_1, \textbf{x}_2, ..., \textbf{x}_N\}$ be the training set. Given a base model $F_i$, a training example $\textbf{x}_i$ is considered a negative sample with respect to $F_i$, if $F_i$ makes wrong prediction on $\textbf{x}_i$. Given a base model $F_i$, the output of $F_i$ on $\mathcal{D}$ is a vector of binary values, denoted as $\boldsymbol{\omega({F_{i})}} = [\omega_{i, 1}, \omega_{i, 2}, ..., \omega_{i, N}]^{T}$, and $\omega_{i, k} = 1$ if $F_i$ predicts $\textbf{x}_k$ correctly, otherwise, $\omega_{i, k} = 0$. We can obtain the set of negative samples based on $\mathcal{D}$ and $\boldsymbol{\omega({F_{i})}}$ for each base model $F_i$.

\subsection{Pairwise Diversity Metrics}
\noindent For pairwise diversity metrics, they are calculated based on a pair of models. Table~\ref{table:relationship-classifier-pair} shows the relationship between a pair of models $F_i$ and $F_j$. For a labeled sample $\textbf{x}_k$, we have four different types of prediction results, such as both $F_i$ and $F_j$ make correct or wrong predictions and either $F_i$ or $F_j$ makes correct predictions. Correspondingly, we can count the number of samples in the four different types, that is $N^{ab}$, which represents the number of elements $\textbf{x}_k \in \textbf{D}$, such that $\omega_{i,k}=a$ and $\omega_{j,k}=b$.

\begin{table}[h!]
\centering
\caption{Relationship Between a Pair of Models}
\label{table:relationship-classifier-pair}
\small
\begin{tabular}{ccc}
\hline
& $F_j$ correct (1) & $F_j$ wrong (0) \\ \hline
$F_i$ correct (1) & $N^{11}$  &  $N^{10}$ \\ \hline
$F_i$ wrong (0) & $N^{01}$ & $N^{00}$  \\ \hline
\multicolumn{2}{c}{$N = N^{00} + N^{01} + N^{10} + N^{11}$} &
\end{tabular}
\end{table}

\subsubsection{Cohen's Kappa (CK)}
The Cohen's Kappa measures the diversity between the two models $F_i, F_j$ from the perspective of agreement~\cite{cohenskappa,diversityaccuracy}. A lower Cohen's Kappa value implies lower agreement and higher diversity. Formula~\ref{formula:cohens-kappa} shows the definition of the Cohen's Kappa ($\kappa_{ij}$) between the two models $F_i, F_j$. The value for the Cohen's Kappa ranges from -1 to 1 with 0 representing the amount of agreement of random chances.~\cite{cohenskappa}
\begin{equation}
\footnotesize
\kappa_{ij} = \frac{2(N^{11}N^{00} - N^{01}N^{10})}{(N^{11}+N^{10})(N^{01}+N^{00})+(N^{11}+N^{01})(N^{10}+N^{00})}
\label{formula:cohens-kappa}
\end{equation}

\subsubsection{Q Statistics (QS)}
The Q statistics~\cite{qstatistics} is defined as $QS_{ij}$ in Formula~\ref{formula:q-statistic} between two models $F_i, F_j$. $QS_{ij}$ varies between -1 and 1. When the models $F_i, F_j$ are statistically independent, the expected $QS_{ij}$ is 0. If the two models tend to recognize the same object similarly, $QS_{ij}$ will have positive value. While two diverse models, recognizing the same object very differently, will render a small or even negative $QS_{ij}$ value.
\begin{equation}
\footnotesize
QS_{ij} = \frac{N^{11}N^{00} - N^{01}N^{10}}{N^{11}N^{00} + N^{01}N^{10}}
\label{formula:q-statistic}
\end{equation}

\subsubsection{Binary Disagreement (BD)}
The binary disagreement~\cite{binarydisagreement,diversityaccuracy} is the ratio between (i) the number of samples on which one model is correct while the other one is wrong to (ii) the total number of samples predicted by the two models $F_i$ and $F_j$ as Formula~\ref{formula:binary-disagreement} shows. 
\begin{equation}
\footnotesize
\theta_{ij} = \frac{N^{01} + N^{10}}{N^{11} + N^{10} + N^{01}+N^{00}}
\label{formula:binary-disagreement}
\end{equation}

For an ensemble team of $S$ models, as recommended by~\cite{diversityaccuracy}, we calculate the averaged metric value over all pair of models as Formula~\ref{formula:averaged-q-statistic} shows, where Q represents a pair-wise diversity metric.

\begin{equation}
\footnotesize
Q = \frac{2}{S(S-1)}\sum_{i=1}^{S-1}\sum_{j=i+1}^{S}Q_{ij}
\label{formula:averaged-q-statistic}
\end{equation}

\subsection{Non-pairwise Diversity Metrics}
\noindent Non-pairwise diversity metrics are primarily designed for a team of $M$ models ($M \ge 3$). In an ensemble team of $S$ models, we use $l(\textbf{x}_k)$ to denote the number of models that correctly learn the class label of $\textbf{x}_k$, i.e., $l(\textbf{x}_k) = \sum_{i=1}^{S} \omega_{ik}$.

\subsubsection{Fleiss' Kappa (FK)}
Similar to Cohen's Kappa, the Fleiss' Kappa~\cite{fleisskappa} also measures the diversity from the perspective of agreement. But it is directly calculated from a team of more than 2 models as Formula~\ref{formula:fleiss-kappa} shows, where $\bar{p}$ is the average classification accuracy for the ensemble team and $\kappa$ is not obtained by simply averaging Cohen's Kappa ($\kappa_{ij}$).
\begin{equation}
\footnotesize
\begin{aligned}
\bar{p} &= \frac{1}{NS} \sum_{k=1}^{N} \sum_{i=1}^{S} \omega_{i,k} \\
\kappa &= 1 - \frac{\frac{1}{S}\sum_{k=1}^{N}l(\textbf{x}_k)(S-l(\textbf{x}_k)}{N(S-1)\bar{p}(1-\bar{p})}
\end{aligned}
\label{formula:fleiss-kappa}
\end{equation}

\subsubsection{Kohavi-Wolpert Variance (KW)}
Kohavi-Wolpert Variance is derived by~\cite{diversityaccuracy} to measure the variability of the predicted class label for the sample $\textbf{x}$ with the team of models $F_1, F_2, ..., F_S$ as Formula~\ref{kw-variance} shows. Higher value of KW variance indicates higher model diversity of the team.

\begin{equation}
\footnotesize
KW = \frac{1}{NS^2}\sum_{k=1}^N l(\textbf{x}_k)(S-l(\textbf{x}_k))
\label{kw-variance}
\end{equation}

\subsubsection{Generalized Diversity (GD)}
The generalized diversity~\cite{generalizeddiversity} metric is given as Formula~\ref{formula:gd}.
$Y$ is a random variable, representing the proportion of models (out of $S$) that fail to recognize a random sample $\textbf{x}$. The probability of $Y=\frac{i}{S}$ is denoted as $p_i$, i.e., the probability of $i$ (out of $S$) models recognizing a randomly chosen sample $\textbf{x}$ incorrectly. $p(1)$ represented the expected probability of one randomly picked model failing while $p(2)$ denotes the expected probability of both two randomly picked models failing. $GD$ varies between 0 and 1. The maximum diversity (1) occurs when the failure of one model is accompanied by the correct recognition by the other model for two randomly picked models, that is $p(2)=0$. When both two randomly picked models fail, we have $p(1)=p(2)$, corresponding to the minimum diversity, 0.

\begin{equation}
\footnotesize
\begin{aligned}
p(1) &= \sum_{i=1}^{S} \frac{i}{S}p_i \\
p(2) &= \sum_{i=1}^{S} \frac{i(i-1)}{S(S-1)} p_i \\
GD &= 1 - \frac{p(2)}{p(1)}
\end{aligned}
\label{formula:gd}
\end{equation}

\subsection{Normalization of Diversity Scores}
\noindent Different diversity metrics use different scoring mechanisms to measure the ensemble diversity. Some Q-diversity metrics use small values to indicate high diversity and others use large values instead. In order to compare all six diversity metrics in a consistent manner, such that the low Q diversity value corresponds to high ensemble diversity, we normalize the Q-value for BD, KW and GD by applying (1$-$Q-value) as the diversity score. In the rest of the paper, ensembles with low Q-diversity scores will imply high ensemble diversity, which indicates high failure independence and weak correlation when making errors. Therefore, high diversity ensembles hold the potential to improve the prediction (testing) performance with high ensemble accuracy. 

\subsection{Q-diversity Ensemble Selection Algorithms}

\noindent We describe the pseudo code for Q-diversity based ensemble selection in Algorithm~\ref{alg:q-enemble-selection}. The algorithm takes the input of a Q-diversity metric, the candidate ensemble set $EnsSet$ generated for a given pool of $M$ models, and the corresponding set of negative samples, {\em NegSampSet}, generated by using the random negative sampling module, for the given base model pool of $M$ models (see Section~\ref{section:negative-sampling} for details). For a given diversity metric, such as GD, we first calculate the Q diversity score for each candidate ensemble in $EnsSet$ (Line~\ref{alg:q-diversity-start}$\sim$\ref{alg:q-diversity-end}) and store them in the set of diversity measurements $Div$. 
Then we calculate the average (mean) value of these diversity scores (Line~\ref{alg:q-enemble-selection-threshold}) as the diversity threshold. This threshold will be used to select those candidate ensembles with their diversity scores smaller than this threshold, indicating high ensemble diversity, and place them in the set of selected ensembles, denoted by $GEnsSet$. Those ensembles whose diversity scores are higher than this mean threshold will be pruned out. 

\begin{algorithm}[!h]
\caption{Q Ensemble Selection (Mean-threshold)}
\label{alg:q-enemble-selection}
\footnotesize
\begin{algorithmic}[1]
    \Procedure{QEnsSelection}{$NegSampSet, Q, EnsSet$}
    \State \textbf{Input}: $NegSampSet$: negative samples; $Q$ the diversity metric; $EnsSet$: the set of ensemble teams to be considered;
    \State \textbf{Output}: $GEnsSet$: the set of good ensemble teams.
    \State Initialize $GEnsSet=\{\}$, $Div=\{\}$
    \For{$i=1$ to $|EnsSet|$} \label{alg:q-diversity-start}
        \LineComment{calculate the diversity metric $Q$ for $T_i\in EnsSet$}
        \State $q_i = DiversityMetric(Q, T_i, NegSampSet)$
        \State $Div$.append($q_i$)
        \Comment{Store $q_i$ in the diversity measures $Div$}
    \EndFor \label{alg:q-diversity-end}
    \State $\theta(Q)=AverageDiversity(Div)$ \label{alg:q-enemble-selection-threshold}
    \Comment{Calculate diversity threshold}
    \For{$i=1$ to $|EnsSet|$} \label{alg:q-enemble-selection-start}
        \If{$q_i < \theta(Q)$}
            \State $GEnsSet$.add($T_i$)
            \Comment{add qualified $T_i$}
        \EndIf
    \EndFor \label{alg:q-enemble-selection-end}
    \State \Return $GEnsSet$
    \EndProcedure
\end{algorithmic}
\end{algorithm}

\begin{figure*}[h!]
\centering
    \subfloat[\small{Any Model (All Teams, CK)}]{
    \centering
    \includegraphics[width=0.33\textwidth]{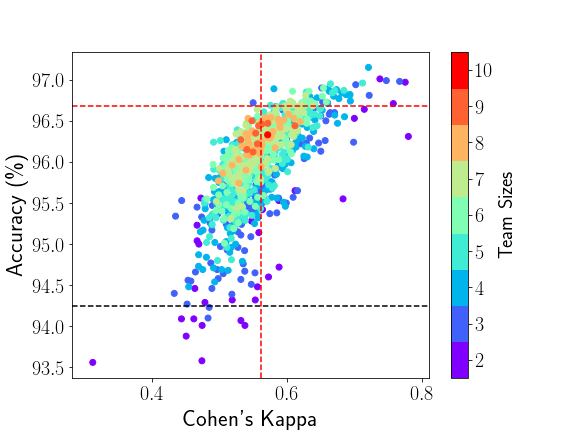}
    \label{fig:cifar10-ck-acc-any-error}
    }
    \subfloat[\small{Any Model (All Teams, GD)}]{
    \centering
    \includegraphics[width=0.33\textwidth]{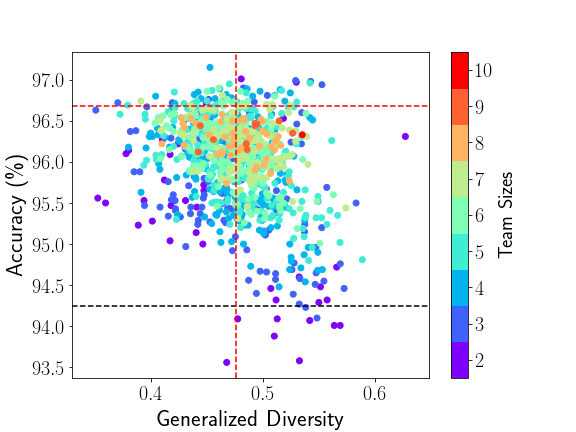}
    \label{fig:cifar10-gd-acc-any-error}
    }
    \subfloat[\small{All Models (All Teams,CK)}]{
    \centering
    \includegraphics[width=0.33\textwidth]{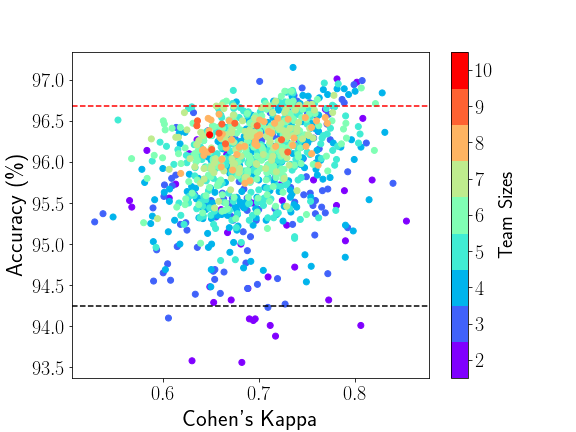}
    \label{fig:cifar10-ck-acc-all-error}
    } \newline
    \subfloat[\small{Target Model $F_3$ (All Teams, CK)}]{
    \centering
    \includegraphics[width=0.33\textwidth]{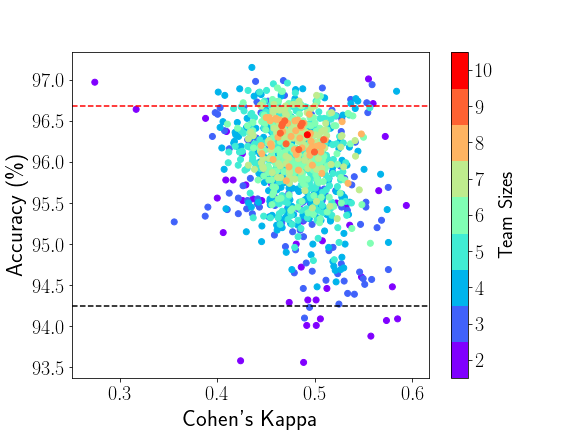}
    \label{fig:cifar10-ck-acc-target-3}
    }
    \subfloat[\small{Target Model $F_3$ ($F_3$, $S=5$, CK)}]{
    \centering
    \includegraphics[width=0.33\textwidth]{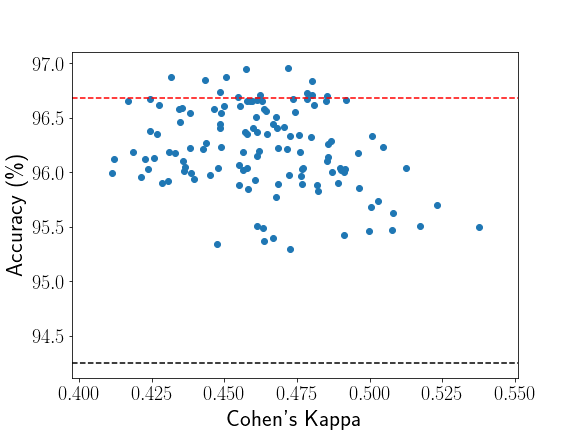}
    \label{fig:cifar10-ck-acc-team-size-5-target-3}
    }
    \subfloat[\small{Target Model $F_3$ ($F_3$, $S=5$, GD}]{
    \centering
    \includegraphics[width=0.33\textwidth]{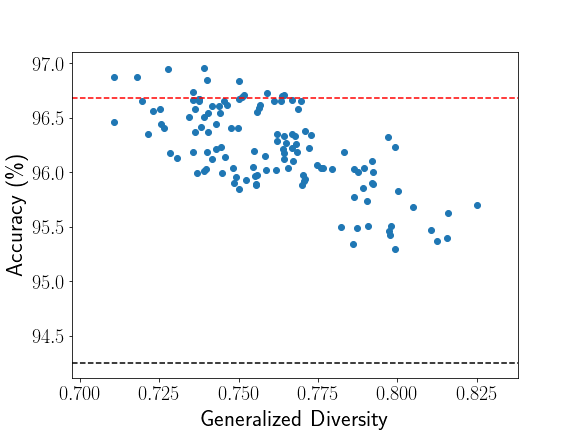}
    \label{fig:cifar10-gd-acc-team-size-5-target-3}
    }
    \caption{The Relationship between Ensemble Accuracy and Diversity under Different Negative Sampling (CIFAR-10)}
    \label{fig:cifar10-negative-sampling}
\end{figure*}

\begin{table}[h!]
\centering
\caption{Ensemble Selection by Q Metrics (ImageNet)}
\label{table:q-ensemble-imagenet}
\scalebox{0.85}{
\small
\begin{tabular}{|c|c|c|c|c|}
\hline
Methods & \#EnsSet & \#GEnsSet & \begin{tabular}[c]{@{}c@{}}Ensemble \\ Acc Range (\%)\end{tabular} & \begin{tabular}[c]{@{}c@{}}Ensemble\\ Acc Avg (\%)\end{tabular} \\ \hline
Baseline & 1013 & 1013 & 61.39$\sim$80.77 & 78.72 \\ \hline
Q-CK\textless{}0.500 & 1013 & 555 & 61.39$\sim$80.50 & 78.16 \\ \hline
Q-QS\textless{}0.697 & 1013 & 483 & 61.39$\sim$80.54 & 78.18 \\ \hline
Q-BD\textless{}0.686 & 1013 & 554 & 61.39$\sim$80.54 & 78.24 \\ \hline
Q-FK\textless{}0.499 & 1013 & 553 & 61.39$\sim$80.50 & 78.15 \\ \hline
Q-KW\textless{}0.878 & 1013 & 647 & \textbf{68.72}$\sim$80.56 & 78.71 \\ \hline
Q-GD\textless{}0.665 & 1013 & 530 & \textbf{70.79}$\sim$80.60 & \textbf{78.77} \\ \hline
\end{tabular}
}
\end{table}

Table~\ref{table:q-ensemble-imagenet} shows the measurement of ensemble selection algorithms using the six Q-diversity metrics on ImageNet. All six algorithms prune out almost 50\% of the candidate ensembles by reducing the total number of selected ensembles to 483$\sim$647 instead of 1013 in $EnsSet$. The good news is that the Q-GD metric and the Q-KW metric based ensemble selection algorithms improve the lower bound of the accuracy range of the selected ensembles from 61.39\% of the baseline to 70.79\% and 68.72\% respectively. Furthermore, the Q-GD algorithm also improves the average ensemble accuracy of the selected ensembles to 78.77\% over the baseline approach of 78.72\%. 
However, there are several inherent problems of using Q-diversity metrics to select good ensembles and prune out bad ensembles. First, five out of six algorithms result in slightly worse accuracy range and average ensemble accuracy for the selected ensembles with marginally lower upper bound and lower average accuracy compared to those of the baseline (no pruning over $EnsSet$). These observations motivate us to drill down the set of factors that may impact on the selection and pruning quality. We first identify a number of factors that are worth in-depth analysis, such as the pruning strategies and the methods used to generate negative samples prior to random sampling over the set of negative examples. Then we design and develop new diversity metrics that optimize the six Q-diversity metrics by preserving the strong points of Q-diversity metrics, such as the statistical utilities of using positive and negative samples to evaluate the degree of negative correlation among the member models of an ensemble, and by improving those weak points identified.

\section{Optimizations of Q-Diversity Metrics}
\subsection{FQ-diversity Metrics}
\noindent The first intuition is based on two observations from our ongoing model ensemble approach to mitigating adversarial attacks to well-trained deep learning models~\cite{ensemble-mass, ensemble-bigdata, robustdlensemble}. First, given a target model (i.e., the victim model), when using Kappa-statistics based diversity to measure diversity of an ensemble team, the negative samples are adversarial examples generated over the target victim model. Second, using the target model based negative samples, we see the comparison among those ensemble teams of equal size using Kappa scores to be more meaningful. Put differently, ensembles, which have equal number of member models and share the same target model, tend to show a clear correlation between ensemble diversity and ensemble accuracy, at the same time, such ensembles also deliver higher robustness to protect the target (victim) model against adversarial perturbed inputs (malicious negative samples). This in-depth analysis combined with extensive empirical measurements motives our development of new diversity metrics~\cite{ensemblebench}, such as FQ-diversity metrics. Our extensive experiments with benchmark datasets show that the new diversity metrics significantly outperform the Q-diversity based ensemble selection algorithms. 
Due to the space limit, we below give an informal overview of our FQ-diversity metrics, focusing on (i) the focal model concept and focal model driven negative sampling, and (ii) the focal model driven diversity computation with fixed model size, and the focal model driven selection and pruning by learning the focal model based threshold.

\subsection{Revisiting Negative Sampling Methods with Focal Model} \label{section:negative-sampling}
\noindent Negative samples are used to calculate the Q-diversity scores for ensembles. 
Given a pool of $M$ base models, negative samples are typically obtained by the union of all $M$ sets of negative examples identified from each base model. We coin this baseline approach as the {\em any\_model} method. For the sake of comparison, we also create the intersection of all these $M$ sets of negative examples, one from each of the $M$ models. We coin this as the {\em all\_models} method. Our random negative sampling module will randomly select a small subset of fixed size, denoted as $NegSampSet$. We will use this randomly selected subset of negative samples to evaluate the Q-diversity scores for every ensemble in the candidate ensemble set $EnsSet$. Figure~\ref{fig:cifar10-ck-acc-any-error}$\sim$\ref{fig:cifar10-ck-acc-all-error} show the relationship between ensemble diversity and ensemble accuracy for Q-CK (any\_model), Q-GD (any\_model) and Q-CK(all\_models) respectively. This set of experiments were conducted on a pool of $M=10$ base models on CIFAR-10. Thus there are 1013 total number of candidate ensembles with varying size S ($2\leq S\leq M$). Different colors indicate different team sizes by the right color diagram. The red and black horizontal dashed lines mark the maximum single model accuracy 96.68\% and the average accuracy of the base model pool 94.25\%. The red vertical dashed lines on Figure~\ref{fig:cifar10-ck-acc-any-error} for Q-CK and~\ref{fig:cifar10-gd-acc-any-error} for Q-GD mark the mean threshold for Q-diversity ensemble selection for CIFAR-10. 
We make three interesting observations: (1) All the three cases show no clear correlation between ensemble diversity and ensemble accuracy. (2) For the same negative sampling method {\em any\_model}, Q-CK and Q-GD show different correlation patterns. Also the mean diversity threshold is ineffective for both Q-diversity metrics. (3) As expected, the Q-CK with the {\em all\_models} negative samples (Figure~\ref{fig:cifar10-ck-acc-all-error}) is the worst, compared to Q-CK and Q-GD using random negative samples from the {\em any\_model} method. These observations indicate the non-trivial relationship between negative sampling and ensemble diversity. Next, we introduce the focal model based negative sampling method, coined as {\em focal\_model} method. Concretely, for ensembles of a fixed size $S$, say $S$=5, we compute the CK diversity score for all ensemble models of size $S$ as follows: by considering each of the $M$ models, say $F_i$ ($0\leq i\leq M-1$), as the focal model (similar to the target model in our ensembles for adversarial learning), we find all ensembles of size S in the candidate set $EnsSet$, which has $F_i$ as a member model. We compute their diversity value for each of the six Q-diversity metrics. Figure~\ref{fig:cifar10-ck-acc-target-3} shows all 1013 ensembles with their FQ-CK values on x-axis and their ensemble accuracy on y-axis. Comparing with Figure~\ref{fig:cifar10-ck-acc-any-error} or Figure~\ref{fig:cifar10-ck-acc-all-error}, we see the focal model based negative sampling is encouraging in terms of improving the quality of the correlation between ensemble diversity and ensemble accuracy. 

\subsection{Focal Model based Diversity Ensemble Selection}
\noindent Now we illustrate the second design principle: introducing ensemble diversity computation and ensemble selection by combining focal model based negative sampling and threshold learning over fixed size ensembles. 
We present the focal model $F_3$ with FQ-CK computed over ensembles of fixed size S=5 in Figure~\ref{fig:cifar10-ck-acc-team-size-5-target-3} and the same focal model $F_3$ with FQ-GD computed over the same set of ensembles of fixed team size S=5 in Figure~\ref{fig:cifar10-gd-acc-team-size-5-target-3}.  
Both of these figures show a much improved relationship between ensemble diversity and ensemble accuracy. However, using the averaging (mean) threshold for selecting good ensembles are clearly inadequate. This motivates us to introduce an initial-centroid optimized k-means clustering to obtain a binary partitioning: one to keep and one to prune out. Based on the boundary of two clusters, we compute the diversity cut-off reference point for each of the six FQ-diversity metrics as the ensemble selection rules. This ensemble selection rule learning process is iteratively performed for all term sizes from $S=2, \dots, M$. For example, given GD metric, the FQ-GD diversity ensemble selection learning will output all the rules related to FQ-GD, which will then be used for selecting high quality ensembles from those to keep and avoid those ensembles that have low ensemble diversity and may result in low ensemble accuracy with high probability (confidence). 

\begin{table*}[h!]
\centering
\caption{10 Examples of Ensembles on CIFAR-10}
\label{table:ensemble-example-cifar10}
\scalebox{0.9}{
\small
\begin{tabular}{|c|c|c|c|c|c|c|c|c|c|c|}
\hline
Ensemble   Team & 0123 & 013 & 012349 & 01238 & 123 & 1234 & 0123456789 & 137 & 01468 & 136789 \\ \hline
Ensemble   Accuracy & \textbf{97.15} & \textbf{96.99} & \textbf{96.90} & \textbf{96.87} & \textbf{96.81} & \textbf{96.63} & \textcolor{red}{96.33} & \textbf{96.31} & \textcolor{red}{96.16} & \textcolor{red}{95.89} \\ \hline
Highest   Member Acc & 96.68 & 96.68 & 96.68 & 96.68 & 96.23 & 96.23 & 96.68 & 96.21 & 96.68 & 96.21 \\ \hline
Highest   Member Model & 0 & 0 & 0 & 0 & 2 & 2 & 0 & 3 & 0 & 3 \\ \hline
Acc   Improvement & \textbf{0.47} & \textbf{0.31} & \textbf{0.22} & \textbf{0.19} & \textbf{0.58} & \textbf{0.40} & \textcolor{red}{-0.35} & \textbf{0.10} & \textcolor{red}{-0.52} & \textcolor{red}{-0.32} \\ \hline
Diversity Measurement (FQ-GD) & 0.0371 & 0.0771 & 0.0522 & 0.1429 & 0.1703 & 0.1264 & 0.6979 & 0.3081 & 0.4291 & 0.5423 \\ \hline
\end{tabular}
}
\end{table*}

\begin{table*}[h!]
\centering
\caption{10 Examples of Ensembles on ImageNet}
\label{table:ensemble-example-imagenet}
\scalebox{0.9}{
\small
\begin{tabular}{|c|c|c|c|c|c|c|c|c|c|c|}
\hline
Ensemble   Team & 12345 & 2345 & 1234 & 124 & 0123456789 & 123467 & 128 & 1289 & 568 & 047 \\ \hline
Ensemble   Accuracy & \textbf{80.77} & \textbf{80.70} & \textbf{80.29} & \textbf{79.84} & \textbf{79.82} & \textbf{79.4}5 & \textbf{78.67} & \textbf{78.62} & \textcolor{red}{77.75} & \textcolor{red}{75.07} \\ \hline
Highest   Member Acc & 78.25 & 78.25 & 77.4 & 77.25 & 78.25 & 77.4 & 77.15 & 77.15 & 78.25 & 77.25 \\ \hline
Highest   Member Model & 5 & 5 & 3 & 4 & 5 & 3 & 1 & 1 & 5 & 4 \\ \hline
Acc   Improvement & \textbf{2.52} & \textbf{2.45} & \textbf{2.89} & \textbf{2.59} & \textbf{1.57} & \textbf{2.05} & \textbf{1.52} & \textbf{1.47} & \textcolor{red}{-0.50} & \textcolor{red}{-2.18} \\ \hline
Diversity Measurement (FQ-GD) & 0 & 0 & 0.0500 & 0.0984 & 0.8459 & 0.2143 & 0.3159 & 0.3475 & 0.3757 & 0.8377 \\ \hline
\end{tabular}
}
\end{table*}

Table~\ref{table:ensemble-example-cifar10} and Table~\ref{table:ensemble-example-imagenet} illustrate the advantages of FQ-diversity metrics by providing 10 example ensembles and the quantitative measures of their ensemble performance on CIFAR-10 and ImageNet respectively. For both benchmark datasets, their respective base model pools of 10 base models are given in Table~\ref{table:base-model-pools}, where the max accuracy of the base models for CIFAR-10 and ImageNet belong to $F_0$ and $F_5$ respectively. In these two sets of experiments, we include the FQ-GD diversity score for each example ensemble. We highlight three interesting observations.
{\it First}, these FQ-GD selected ensembles are high quality ensembles. For example, the first five ensemble teams outperform $F_0$, the base model with the max accuracy (96.68\%) for CIFAR-10. Although the first four all contain $F_0$ (\verb|0123|, \verb|013|, \verb|012349|, \verb|01238|), the ensemble \verb|123| delivers ensemble accuracy of 96.81\% without $F_0$ as its member model.
Similarly, the first eight ensembles all outperform $F_5$, the base model with the max accuracy (78.25\%) for ImageNet. Only three of them (\verb|12345|, \verb|2345|, \verb|0123456789|) has $F_5$ as a member model. 
Their diversity measurements show very small value (\textless{} 0.35), indicating very high ensemble diversity, except the ensemble of size $M$. 
{\it Second}, 
having the top performing base model (max accuracy) as a member model of an ensemble may not guarantee that the ensemble accuracy is on par to or higher than the max accuracy of its member models. For instance, the ensemble team \verb|01468| contains $F_0$, the max accuracy base model for CIFAR-10 and the ensemble \verb|568| has $F_5$, the max accuracy base model, as its member model for ImageNet, but \verb|01468| has ensemble accuracy below 96.68\% on CIFAR-10 and \verb|568| has ensemble accuracy below 78.25\% on ImageNet. It is also interesting to see that their FQ-GD diversity scores are high, 
0.4291 for \verb|01468| and 0.3757 for \verb|568|.
{\it Third}, the ensemble of size $M$, i.e., including all $M$ base models, may not guarantee the top ensemble accuracy, compared to other ensembles of smaller team sizes. For example, on CIFAR-10, the ensemble team \verb|0123456789| only achieved 96.33\% accuracy, which is lower than the first six ensembles of smaller team sizes, and also lower than the max accuracy (96.68\%) of its top performing member model $F_0$. Similarly, for ImageNet, the ensemble team \verb|0123456789| achieved 79.82\% accuracy, which is lower than the first four ensembles, each of which has a smaller team and a subset of the $M$ base models, yet higher ensemble accuracy with the ensemble team \verb|12345| achieving ensemble accuracy of 80.77\%. 

A complete experimental evaluation on CIFAR-10 and ImageNet will be reported in Section~\ref{section:experiments}. In addition to FQ-diversity metrics, EnsembleBench also implemented other new diversity metrics, such as HQ-diversity metrics and the alpha filter~\cite{ensemblebench}, which preemptively prunes out low diversity ensembles during the candidate ensemble formation process. EnsembleBench also supports diversity metric fusion (EQ), which combines the high diversity ensembles selected by top performing FQ-diversity metrics to generate and recommend an elite set of high diversity ensembles. For a given learning task, dataset and pool of base models, EnsembleBench can perform diversity evaluation in seconds and output recommended high diversity ensembles. These diversity measurements and corresponding calculated diversity-based selection rules will be stored for serving future ensemble queries.

\section{Ensemble Consensus Algorithms} \label{section:consensus-voting}
\noindent Given an ensemble of size $S$, an ensemble consensus algorithm will combine the prediction results of these $S$ member models. Representative ensemble consensus algorithms include soft voting, plurality voting and majority voting. For a sample $\textbf{x}_k$, a probabilistic model $F_i$, such as a neural network, will output the prediction probability vector, $P_i(\textbf{x}_k)$, for all $C$ classes, and the probability for each class $c_j$ is $P_i(c_j|\textbf{x}_k)$. The prediction result on the sample $\textbf{x}_k$ is the class with the highest probability, i.e., $pred_k = argmax_j P(c_j|\textbf{x}_k)$. For neural networks, the probability vector can be obtained with the output of the last softmax layer.

\textbf{i. Soft Voting (Model Averaging)}
Soft voting predicts the class label based on the average prediction probabilities as Formula~\ref{formula:soft-voting} shows, where $P_E(c_j | \textbf{x}_k)$ is the combined prediction probability to give the predicted class.

\begin{equation}
\begin{aligned}
\footnotesize
P_E(c_j | \textbf{x}_k) = \frac{1}{S} \sum_{i=1}^{S} P_i(c_j | \textbf{x}_k) \\
pred_k = argmax_j P_E(c_j | \textbf{x}_k)
\end{aligned}
\label{formula:soft-voting}
\end{equation}

\textbf{ii. Majority Voting} For a sample $\textbf{x}_k$, majority voting will count all the votes (predicted labels ($pred_k$)) by all the member models in an ensemble. The class received the majority of the votes (more than half) will be selected as the ensemble prediction label.

\textbf{iii. Plurality Voting} Plurality voting is very similar to majority voting. It also counts all the votes for predicted classes by all member models. The class received the highest number of votes will be chosen as the ensemble prediction label. Note that majority voting is typically confused with plurality voting while plurality voting does not require majority for giving the final prediction.\cite{voting-comparison}

\textbf{iv. Boosting Voting} The above consensus algorithms assume equal weights of member models in consensus analysis of an ensemble. An alternative approach is to design weighted consensus algorithms. For example, learn to combine algorithms will use machine learning algorithms to train a model over the training dataset, each training sample is a collection of votes from member models of an ensemble. The training will learn a model that can identify a proper weight for each member model. We implement a boosting voting algorithm to learn the weights for member models, and such weighted consensus will be used for ensemble predictions. Algorithm~\ref{alg:boosting-voting} shows the pseudo code for learning the weights for member models. For an ensemble team $F$, the training data $X$ with label $Y$, and $\gamma$ penalty parameter ($\gamma < 0$ and $\gamma=-0.01$ by default). The weights $w_i$ will be first initialized with equal weights $\frac{1}{S}$ (Line~\ref{alg:boosting-voting-init}). Then for each training sample $X_k$, for each member model, if the model $F_i$ makes the wrong prediction, its weight will be penalized by multiplying $e^{\gamma} < 1$ similar to boosting algorithms (Line~\ref{alg:boosting-voting-adjust-w-start}$\sim$\ref{alg:boosting-voting-adjust-w-end}). After finishing adjusting the weights for all member model, the weights will be normalized to ensure the sum is still 1. Using the returned member model weights $W$, we then follow the weighted soft voting as Formula~\ref{formula:boosting-voting} shows to output the prediction results.

\begin{algorithm}
\caption{Boosting Voting}
\label{alg:boosting-voting}
\footnotesize
\begin{algorithmic}[1]
    \Procedure{boostingVoting}{$F, X, Y, \gamma$}
    \State \textbf{input}: $F$ the ensemble team of $S$ models; $X$ training samples; the corresponding ground-truth labels $Y$, $\gamma$ the penalty parameter ($\gamma$\textless{} 0).
    \State \textbf{output}: $W$ the weights for all member models
    \State Initialize $W=\{w_1, w_2, ..., w_S\}$ with $\frac{1}{S}$ \label{alg:boosting-voting-init}
    \Comment{initialize the weights}
    \For{$k=1$ to $|X|$}
        \Comment{test on every testing sample}
        \For{$i = 1$ to $|F|$} \label{alg:boosting-voting-adjust-w-start}
        \Comment{accumulate weights from each model}
            \State $y = argmax_j P_i(c_j | X_k)$
            \If{$y != Y_k$}
            \Comment{apply penalty for the wrong prediction}
                \State $w_i = w_i * e^{\gamma}$
                \Comment{$\gamma < 0$, we have $e^{\gamma}<1$}
            \EndIf
        \EndFor \label{alg:boosting-voting-adjust-w-end}
        \State normalize $W$ s.t. $sum(W)=1$ \label{alg:boosting-voting-normalize}
    \EndFor
    \State \Return $W$
    \EndProcedure
\end{algorithmic}
\end{algorithm}

\begin{equation}
\begin{aligned}
\footnotesize
P_E(c_j | \textbf{x}_k) = \frac{1}{S} \sum_{i=1}^{S} w_i P_i(c_j | \textbf{x}_k) \\
pred_k = argmax_j P_E(c_j | \textbf{x}_k)
\end{aligned}
\label{formula:boosting-voting}
\end{equation}


\section{Experimental Analysis} \label{section:experiments}
\noindent We perform extensive experiments on two benchmark datasets (CIFAR-10~\cite{cifar10-100} and ImageNet~\cite{ILSVRC}) to evaluate different high diversity ensemble selection algorithms with EnsembleBench. All experiments are conducted on a GPU server with an Intel Xeon E5-1620 CPU and Nvidia GeForce GTX 1080 Ti (11 GB) GPU, installed with Ubuntu 16.04 LTS and CUDA 8.0. For each dataset, Table~\ref{table:base-model-pools} lists the 10 base models.

\begin{table}[h]
\centering
\caption{Base Model Pools for Two Benchmark Datasets}
\label{table:base-model-pools}
\scalebox{0.83}{
\small
\begin{tabular}{|c|cc|cc|}
\hline
\multirow{2}{*}{Dataset} & \multicolumn{2}{c|}{CIFAR-10} & \multicolumn{2}{c|}{ImageNet} \\ \cline{2-5}
 &\multicolumn{2}{c|}{10,000 testing samples} & \multicolumn{2}{c|}{50,000 testing samples} \\ \hline
Number & Models & Accuracy (\%) & Models & Accuracy (\%)  \\ \hline
0 & DenseNet190 & 96.68 & AlexNet & 56.63 \\ \hline
1 & DenseNet100 & 95.46 & DenseNet & 77.15 \\ \hline
2 & ResNeXt & 96.23 & EfficientNet-B0 & 75.80 \\ \hline
3 & WRN & 96.21 & ResNeXt50 & 77.40\\ \hline
4 & VGG19 & 93.34 & Inception3 & 77.25 \\ \hline
5 & ResNet20 & 91.73 & ResNet152 & 78.25 \\ \hline
6 & ResNet32 & 92.63 & ResNet18 & 69.64 \\ \hline
7 & ResNet44 & 93.10 & SqueezeNet & 58.00 \\ \hline
8 & ResNet56 & 93.39 & VGG16 & 71.63 \\ \hline
9 & ResNet110 & 93.68 & VGG19-BN & 74.22 \\ \hline
MIN & ResNet20 & 91.73 & AlexNet & 56.63 \\ \hline
AVG &  & 94.25 &  & 71.60 \\ \hline
MAX & DenseNet190 & 96.68 & ResNet152 & 78.25 \\ \hline
\end{tabular}
} 
\end{table}

\begin{table*}[h!]
\centering
\caption{The Experimental Comparison of Q/FQ/EQ Ensemble Selection (CIFAR-10)}
\label{table:experimental-comparison-cifar10}
\scalebox{0.80}{
\small
\begin{tabular}{|c|c|c|c|c|c|c|c|c|c|}
\hline
Methods & \#EnsSet & \#GEnsSet & \begin{tabular}[c]{@{}c@{}}Ensemble Acc\\ Range (\%)\end{tabular} & \begin{tabular}[c]{@{}c@{}}Ensemble\\Acc Avg (\%)\end{tabular} & STD & \begin{tabular}[c]{@{}c@{}}\# (Acc \textgreater{}=\\  m\_max)\end{tabular} & \begin{tabular}[c]{@{}c@{}}\% (Acc \textgreater{}=\\ m\_max)\end{tabular} & \begin{tabular}[c]{@{}c@{}}\# (Acc \textgreater{}=\\96.68\% p\_max)\end{tabular} & \begin{tabular}[c]{@{}c@{}}\% (Acc \textgreater{}= \\ 96.68\% p\_max)\end{tabular} \\ \hline
Baseline & 1013 & 1013 & 93.56$\sim$97.15 & 95.99 & 0.5513 & 202 & 19.94\% & 66 & 6.52\% \\ \hline
Q-Ensemble (CK\textless{}0.562) & 1013 & 544 & 93.56$\sim$96.72 & 95.69 & 0.5195 & 79 & 14.52\% & 1 & 0.18\% \\ \hline
Q-Ensemble (QS\textless{}0.515) & 1013 & 516 & 93.56$\sim$96.74 & 95.75 & 0.5374 & 82 & 15.89\% & 4 & 0.78\% \\ \hline
Q-Ensemble (BD\textless{}0.661) & 1013 & 550 & 93.56$\sim$96.72 & 95.71 & 0.5187 & 78 & 14.18\% & 2 & 0.36\% \\ \hline
Q-Ensemble (FK\textless{}0.561) & 1013 & 541 & 93.56$\sim$96.72 & 95.69 & 0.5205 & 79 & 14.60\% & 1 & 0.18\% \\ \hline
Q-Ensemble (KW\textless{}0.868) & 1013 & 586 & 94.27$\sim$96.74 & 95.93 & 0.4433 & 61 & 10.41\% & 5 & 0.85\% \\ \hline
Q-Ensemble (GD\textless{}0.476) & 1013 & 496 & 93.56$\sim$97.15 & 96.14 & 0.4332 & 79 & 15.93\% & 36 & 7.26\% \\ \hline \hline
FQ-Ensemble (BD) & 1013 & 369 & 95.23$\sim$97.15 & 96.37 & 0.3385 & 109 & 29.54\% & 60 & 16.26\% \\ \hline
FQ-Ensemble (KW) & 1013 & 370 & 95.09$\sim$97.15 & 96.36 & 0.3461 & 110 & 29.73\% & 61 & 16.49\% \\ \hline
FQ-Ensemble (GD) & 1013 & 443 & 95.23$\sim$97.15 & 96.38 & 0.3165 & 120 & 27.09\% & 66 & 14.90\% \\ \hline
EQ-Ensemble (BD+KW+GD) & 1013 & 336 & 95.23$\sim$97.15 & 96.40 & 0.3298 & 107 & 31.85\% & 60 & 17.86\% \\ \hline
\end{tabular}
}
\end{table*}

\begin{table*}[h!]
\centering
\caption{The Experimental Comparison of Q/FQ/EQ Ensemble Selection (ImageNet)}
\label{table:experimental-comparison-imagenet}
\scalebox{0.80}{
\small
\begin{tabular}{|c|c|c|c|c|c|c|c|c|c|}
\hline
Methods & \#EnsSet & \#GEnsSet & \begin{tabular}[c]{@{}c@{}}Ensemble Acc\\ Range (\%)\end{tabular} & \begin{tabular}[c]{@{}c@{}}Ensemble \\ Acc Avg (\%) \end{tabular} & STD & \begin{tabular}[c]{@{}c@{}}\# (Acc \textgreater{}=\\  m\_max)\end{tabular} & \begin{tabular}[c]{@{}c@{}}\% (Acc \textgreater{}=\\ m\_max)\end{tabular} & \begin{tabular}[c]{@{}c@{}}\# (Acc \textgreater{}=\\ 78.25\% p\_max)\end{tabular} & \begin{tabular}[c]{@{}c@{}}\% (Acc \textgreater{}=\\ 78.25\% p\_max)\end{tabular} \\ \hline
Baseline & 1013 & 1013 & 61.39$\sim$80.77 & 78.72 & 1.6141 & 884 & 87.27\% & 753 & 74.33\% \\ \hline
Q-Ensemble (CK\textless{}0.500) & 1013 & 555 & 61.39$\sim$80.50 & 78.16 & 1.8317 & 434 & 78.20\% & 338 & 60.90\% \\ \hline
Q-Ensemble (QS\textless{}0.697) & 1013 & 483 & 61.39$\sim$80.54 & 78.18 & 1.9203 & 378 & 78.26\% & 296 & 61.28\% \\ \hline
Q-Ensemble (BD\textless{}0.686) & 1013 & 554 & 61.39$\sim$80.54 & 78.24 & 1.8325 & 438 & 79.06\% & 349 & 63.00\% \\ \hline
Q-Ensemble (FK\textless{}0.499) & 1013 & 553 & 61.39$\sim$80.50 & 78.15 & 1.8376 & 431 & 77.94\% & 336 & 60.76\% \\ \hline
Q-Ensemble (KW\textless{}0.878) & 1013 & 647 & 68.72$\sim$80.56 & 78.71 & 1.3345 & 556 & 85.94\% & 473 & 73.11\% \\ \hline
Q-Ensemble (GD\textless{}0.665) & 1013 & 530 & 70.79$\sim$80.60 & 78.77 & 1.2930 & 454 & 85.66\% & 394 & 74.34\% \\ \hline \hline
FQ-Ensemble (BD) & 1013 & 550 & 74.65$\sim$80.77 & 79.47 & 0.7837 & 541 & 98.36\% & 510 & 92.73\% \\ \hline
FQ-Ensemble (KW) & 1013 & 563 & 74.65$\sim$80.77 & 79.45 & 0.7840 & 554 & 98.40\% & 521 & 92.54\% \\ \hline
FQ-Ensemble (GD) & 1013 & 539 & 75.27$\sim$80.77 & 79.51 & 0.7394 & 531 & 98.52\% & 504 & 93.51\% \\ \hline
EQ-Ensemble (BD+KW+GD) & 1013 & 512 & 75.27$\sim$80.77 & 79.52 & 0.7468 & 504 & 98.44\% & 479 & 93.55\% \\ \hline
\end{tabular}
}
\end{table*}

\textbf{CIFAR-10:}
Table~\ref{table:experimental-comparison-cifar10} shows the experimental comparison of using Q/FQ/EQ ensemble selection algorithms in EnsembleBench. The criteria described in Section~\ref{section:ensemble-evaluation} are used to evaluate these ensemble selection algorithms, where the ensemble accuracy range, average ensemble accuracy and the standard deviation of ensemble accuracy of selected ensembles are shown in the 4th$\sim$6th columns. The last 4 columns show the the number and percentage (over \#GEnsSet) of (i) the selected ensembles that have higher accuracy than all its member models (Acc \textgreater{}= m\_max) and (ii) the selected ensembles that have higher accuracy than the best base model in the pool (Acc \textgreater{}= p\_max). We also listed the number of candidate ensembles (\#EnsSet=1013 on 10 base models) and the number of selected ensembles (\#GEnsSet) in the 2nd and 3rd columns. We highlight three interesting observations.
{\it First}, Q ensemble selection algorithms can prune out about half of candidate ensembles and identify 496$\sim$586 ensemble teams out of 1013 candidate ensembles. However, even though some Q metrics, such as Q-KW can improve the ensemble accuracy lower bound to 94.27\%, and Q-GD can improve the average ensemble accuracy from 95.99\% of the baseline to 96.14\%, Q ensemble selection algorithms in general suffer from lower ensemble accuracy upper bound (96.72\%$\sim$96.74\% except Q-GD), and the probability of randomly picking up high quality ensembles among the selected ensembles is also very low, ranging from 10.41\% to 15.93\%.
{\it Second}, we present the top 3 performing FQ metrics, BD, KW and GD. The ensembles selected by these optimized diversity metrics significantly outperformed corresponding Q metrics all measurement criteria. For example, FQ-KW can improve the ensemble accuracy lower bound of selected ensembles from 93.56\% to 95.09\% and still maintain 29.73\% probability of randomly picking up high quality ensembles, which is very high compared to the baseline and Q metrics with less than 19.94\%.
{\it Third}, with diversity metric fusion, the EQ metric, combining the top 3 performing FQ metrics, BD, KW and GD, achieved the best performance. The average ensemble accuracy reaches 96.40\% with the high quality ensembles accounting for 31.85\% of the selected ensembles, the highest among all the ensemble selection algorithms.

\textbf{ImageNet:}
We performed the same set of experiments on ImageNet. The experimental results are shown on Table~\ref{table:experimental-comparison-imagenet}. We found similar observations as CIFAR-10.
{\it First}, some Q metrics, such as Q-KW and Q-GD, can improve the baseline in terms of the ensemble accuracy lower bound or average ensemble accuracy of the selected ensembles. 
{\it Second}, FQ metrics can significantly improve the quality of selected ensembles over the baseline as well as corresponding Q metrics. All three FQ metrics, BD, KW, and GD, significantly improve the ensemble accuracy lower bound to 74.65\% from 61.39\%$\sim$70.79\%. It further demonstrates that FQ metrics can better capture the negative correlation among member models and will be more effective in selecting high quality ensembles.
{\it Third}, the EQ metric, combining the top 3 performing FQ metrics, can further reduce the number of selected ensembles from 539$\sim$550 to 512 among which 504 ensembles (98.44\%) have accuracy higher than all their member models, which indicates a very high chance of picking up the high quality ensembles from the set of high diversity ensembles ($GEnsSet$).

\begin{table}[h]
\centering
\caption{Different Ensemble Consensus Algorithms}
\label{table:ensemble-consensus}
\scalebox{0.95}{
\small
\begin{tabular}{|c|c|c|c|c|c|}
\hline
\multirow{2}{*}{Dataset} & \multirow{2}{*}{Team} & \multicolumn{4}{c|}{Accuracy (\%)} \\ \cline{3-6} 
 &  & Soft & Majority & Plurality & Boosting \\ \hline
\multirow{5}{*}{CIFAR-10} & 0123 & \textbf{97.15} & 95.98 & 96.98 & 97.01 \\ \cline{2-6} 
 & 023 & 96.98 & 96.86 & 96.98 & \textbf{97.06} \\ \cline{2-6} 
 & 01347 & 96.71 & 96.32 & 96.54 & \textbf{97.02} \\ \cline{2-6} 
 & 235 & 96.38 & 96.08 & 96.33 & \textbf{96.70} \\ \cline{2-6} 
 & 1237 & 96.47 & 95.19 & 96.44 & \textbf{96.76} \\ \hline
\multirow{5}{*}{ImageNet} & 12345 & \textbf{80.77} & 78.26 & 80.20 & 80.74 \\ \cline{2-6} 
 & 2345 & \textbf{80.70} & 74.88 & 79.93 & 80.68 \\ \cline{2-6} 
 & 12458 & 80.43 & 77.47 & 79.82 & \textbf{80.54} \\ \cline{2-6} 
 & 1345689 & 80.09 & 76.63 & 79.37 & \textbf{80.26} \\ \cline{2-6} 
 & 13578 & 79.53 & 75.58 & 78.63 & \textbf{79.88} \\ \hline
\end{tabular}
} 
\end{table}

\textbf{Ensemble Consensus Algorithms:}
We further perform a set of experiments to study the impacts of different ensemble consensus algorithms. Table~\ref{table:ensemble-consensus} shows the experimental results of using 4 different ensemble consensus algorithms on the ensembles that identified by the EQ metric on CIFAR-10 and ImageNet. We highlight two interesting observations.
{\it First}, all four ensemble consensus algorithms have reasonable consensus performance, because ensembles of high ensemble diversity still produce high ensemble accuracy when using anyone of the four consensus algorithms.
{\it Second}, soft voting and boosting voting work slightly more effective for the base model pools of the two benchmark datasets used in our experiments, as they tend to produce slightly higher ensemble accuracy than the other two consensus methods. We use boldface to emphasize the highest accuracy for each ensemble team. The highest ensemble accuracy is achieved with either soft voting or boosting voting. In EnsembleBench, the default consensus method is soft voting.

\section{Conclusion}
\noindent 
We have presented EnsembleBench, a general framework for evaluating and recommending high diversity and high accuracy ensembles with three novel features: First, EnsembleBench introduces a set of quantitative metrics for assessing the quality of ensembles and for comparing alternative ensembles constructed for the same learning tasks. Second, EnsembleBench implements a suite of baseline diversity metrics and optimized diversity metrics for identifying and selecting ensembles with high diversity and high quality, making it an effective framework for benchmarking, evaluating and recommending high diversity model ensembles. Third, Four representative ensemble consensus methods are provided in the first release of EnsembleBench, enabling empirical study on the impact of consensus methods on ensemble accuracy. Experimental evaluation on popular benchmark datasets demonstrates the utility and effectiveness of EnsembleBench for promoting high diversity ensembles and boosting the overall performance of selected ensembles.

\section*{Acknowledgment}
This research is partially sponsored by National Science Foundation under NSF 1564097, NSF 2038029, a Cisco grant, and an IBM faculty award. 

\bibliographystyle{IEEEtran}
\bibliography{reference}
%



\end{document}